# FlashSegmentator: High-Throughput TotalSegmentator Inference without Loss of Accuracy

Yosuke Yamagishi[1], Tomohiro Kikuchi[2,3], Shouhei Hanaoka[1,4], Takeharu Yoshikawa[2], Osamu Abe[1,4]

[1] Division of Radiology and Biomedical Engineering, Graduate School of Medicine, The University of Tokyo, Bunkyo-ku, Tokyo, Japan

[2] Department of Computational Diagnostic Radiology and Preventive Medicine, The University of Tokyo Hospital, Tokyo, Japan

[3] Department of Radiology, School of Medicine, Jichi Medical University, Shimotsuke, Tochigi, Japan

[4] Department of Radiology, The University of Tokyo Hospital, Tokyo, Japan

Corresponding author: Yosuke Yamagishi (yamagishi-yosuke0115@g.ecc.u-tokyo.ac.jp)

# Abstract

**Purpose:** To develop, evaluate, and publicly release FlashSegmentator, an open-source inference pipeline that increases the throughput of TotalSegmentator without retraining or loss of segmentation accuracy.

**Materials and Methods:** This retrospective study used public datasets: 73 CT examinations from the TotalSegmentator test split (median age, 65 years; 37 male, 35 female, one unknown), 30 abdominal CT examinations from the Beyond the Cranial Vault (BTCV) dataset, 512 training-split CT examinations for concurrent-process analysis, and 55 MRI examinations. FlashSegmentator adds static-shape compilation, pipelining, batched sliding-window inference, and GPU postprocessing. Throughput was compared with the reference implementation (including its lower-resolution modes) and FastSegmentator on GH200, A100, and T4 GPUs. Dice agreement was tested with two one-sided tests (margin, ±0.01).

**Results:** FlashSegmentator processed 3782, 1810, and 270 examinations per hour on the GH200, A100, and T4 (57.7, 33.9, and 6.4 times the reference implementation and 3.20, 2.50, and 1.22 times compiled FastSegmentator). Dice scores were equivalent for all 117 structures on every GPU (mean difference ≤ $1.2 \times 10^{-5}$; $P < .001$) and at the examination level on BTCV. In contrast, the lower-resolution modes of the reference implementation processed 181 and 213 examinations per hour on the A100 but reduced mean Dice by 0.116 and 0.241. On MRI, throughput increased 94-fold with all 50 structures equivalent. Estimated A100 cloud cost per 10,000 examinations decreased from US $997 to US $29.

**Conclusion:** FlashSegmentator increased the throughput of TotalSegmentator on data center and entry-level GPUs without loss of segmentation accuracy compared with the reference implementation.

# Introduction

Automated whole-body CT segmentation has made it possible to extract quantitative imaging biomarkers from routinely acquired CT examinations. TotalSegmentator, a deep learning model based on the nnU-Net framework, segments more than 100 anatomic structures on CT and has been widely adopted in research (1). Its outputs have been used for the opportunistic assessment of conditions that are not the primary indication for imaging, such as sarcopenia, cardiometabolic risk, osteoporosis, and organ enlargement (2–5), and automated segmentation has been applied to tens of thousands of CT examinations to establish age- and sex-specific reference charts of organ volumes and attenuation (6–9).

Because such measurements require no additional radiation exposure or cost, there is growing interest in applying them to all CT examinations in an institution or to entire imaging archives (3,10).

Applying such tools at this scale, however, is limited by inference time. With the reference implementation, segmenting a single examination takes approximately 1–4 minutes on a high-end consumer GPU (1), so that processing the tens to hundreds of thousands of CT examinations accumulated at one institution can occupy a GPU for weeks to months, particularly with the less powerful GPUs available at many institutions. The same burden recurs whenever a cohort must be reprocessed with an updated model, and prospective use, such as segmenting every examination on the day of acquisition, requires processing that keeps pace with daily scan volume.

Existing approaches to shortening inference either run the model at a coarser resolution, as in the lower-resolution modes of TotalSegmentator, at the cost of segmentation accuracy, or make more efficient use of the GPU. Restricting the task to selected structures is not applicable when whole-body quantification is required. FastSegmentator, an open-source reimplementation of the TotalSegmentator inference pipeline, moves preprocessing and postprocessing to the GPU and reproduces the reference outputs substantially faster (11); however, like the reference implementation, it processes examinations sequentially, one image patch at a time, so that much of the GPU capacity remains unused.

Whether inference-level optimizations alone, without changing the trained model, can substantially increase throughput has not been systematically evaluated. In general-purpose deep learning, batching, overlapping data loading with computation, and compilation are routinely used to improve GPU utilization (12), but medical image segmentation pipelines involve large three-dimensional volumes, resampling to and from a fixed resolution, and an ensemble of networks, so that the benefit of each optimization may depend on the GPU used. Moreover, because any change to the computation can alter the output at the voxel level, an accelerated pipeline must be shown not to reduce accuracy relative to the reference implementation before it can be used in its place.

The purpose of this study was to develop and evaluate FlashSegmentator, an optimized inference pipeline for TotalSegmentator that increases throughput without loss of segmentation accuracy. We compared its throughput with that of the reference implementation, including its lower-resolution modes, and FastSegmentator on three GPU classes, compared segmentation accuracy with that of the reference implementation on two public datasets, and examined its application to the MRI model.

# Materials and Methods

## Study design and data

This retrospective study used only publicly available, deidentified datasets; institutional review board approval was therefore not required.

The following datasets were used. (1) TotalSegmentator dataset version 2.0.1 (1,13): the 89 CT examinations of the official test split, with reference standard segmentations of the 117 structures of the total task created by the dataset authors with model assistance and human review. Sixteen examinations, whose image headers contained direction cosines that were not exactly orthonormal and for which the reference implementation and FastSegmentator failed when writing the output, were excluded from all comparisons, leaving 73 examinations. (2) Multi-Atlas Labeling Beyond the Cranial Vault (BTCV) abdomen dataset (14): 30 training CT examinations with manual segmentations of 13 abdominal structures, to which the corresponding TotalSegmentator labels were mapped. (3) For the analysis of concurrent processes only, 512 CT examinations randomly selected from the training split of the TotalSegmentator dataset (after the same exclusion as above); reference standard segmentations were not used. (4) TotalSegmentator MRI dataset version 2.0.0 (15,16): the 55 MRI examinations of the official test split (50 structures of the total_mr task); six examinations with empty reference standard masks were excluded from the Dice analysis.

## Segmentation model

All implementations used the same pretrained TotalSegmentator total model for CT (five nnU-Net (17) networks at 1.5-mm isotropic resolution) with unchanged weights and identical inference settings (patch size 128 × 128 × 128 voxels, step size 0.8, Gaussian weighting, no test-time mirroring, fp16 mixed precision). For MRI, the pretrained total_mr model (two networks) was used in the same way.

## Implementations compared

The reference implementation was TotalSegmentator version 2.18.0 (https://github.com/wasserth/TotalSegmentator), run through its command-line interface with one call per examination, as in typical use. Because it has no built-in option for parallel processing, it was also run as multiple concurrent calls (xargs -P). Its lower-resolution modes, --fast (3 mm) and --fastest (6 mm), were also evaluated. FastSegmentator (https://github.com/bowang-lab/FastSegmentator) loads the models once and processes examinations sequentially; it was evaluated with and without its built-in compilation option.

FlashSegmentator (this study) was built on FastSegmentator, with preprocessing, network, and postprocessing semantics unchanged, and incorporated six modifications (Figure 1):

1. Static-shape compilation: each network was compiled once with PyTorch 2 (torch.compile) (12), which fuses its operations into optimized GPU code, for a fixed input shape (static shape); incomplete batches were zero-padded.
2. cuDNN autotuning: the fastest convolution algorithm for each layer was selected by benchmarking at run time.
3. Pipelining: examinations were processed in groups of four; while the GPU processed the current group, CPU threads read and resampled the next group and wrote the previous one.
4. Batched sliding window: image patches from all examinations of a group were pooled and passed through each network in batches.
5. GPU postprocessing: conversion of network outputs to labels and merging of the five label maps were performed on the GPU.
6. Output writing: the header read with the input was reused instead of re-reading the input image, and 8-bit labels were written.

The batch size was 8 on the GH200 and 4 on the A100 and T4. As an option, FlashSegmentator can crop the image to the bounding box of the body before inference, which reduced the number of patches by approximately 25% but makes the output not identical to that of the reference implementation; it was therefore evaluated separately (details in Supplemental Material).

On the GH200, the contribution of each modification was assessed by adding the modifications cumulatively. Alternatives not adopted (TensorRT, ONNX Runtime, network fusion, channels-last layout) were compared with the compiled PyTorch networks by network forward time on the GH200.

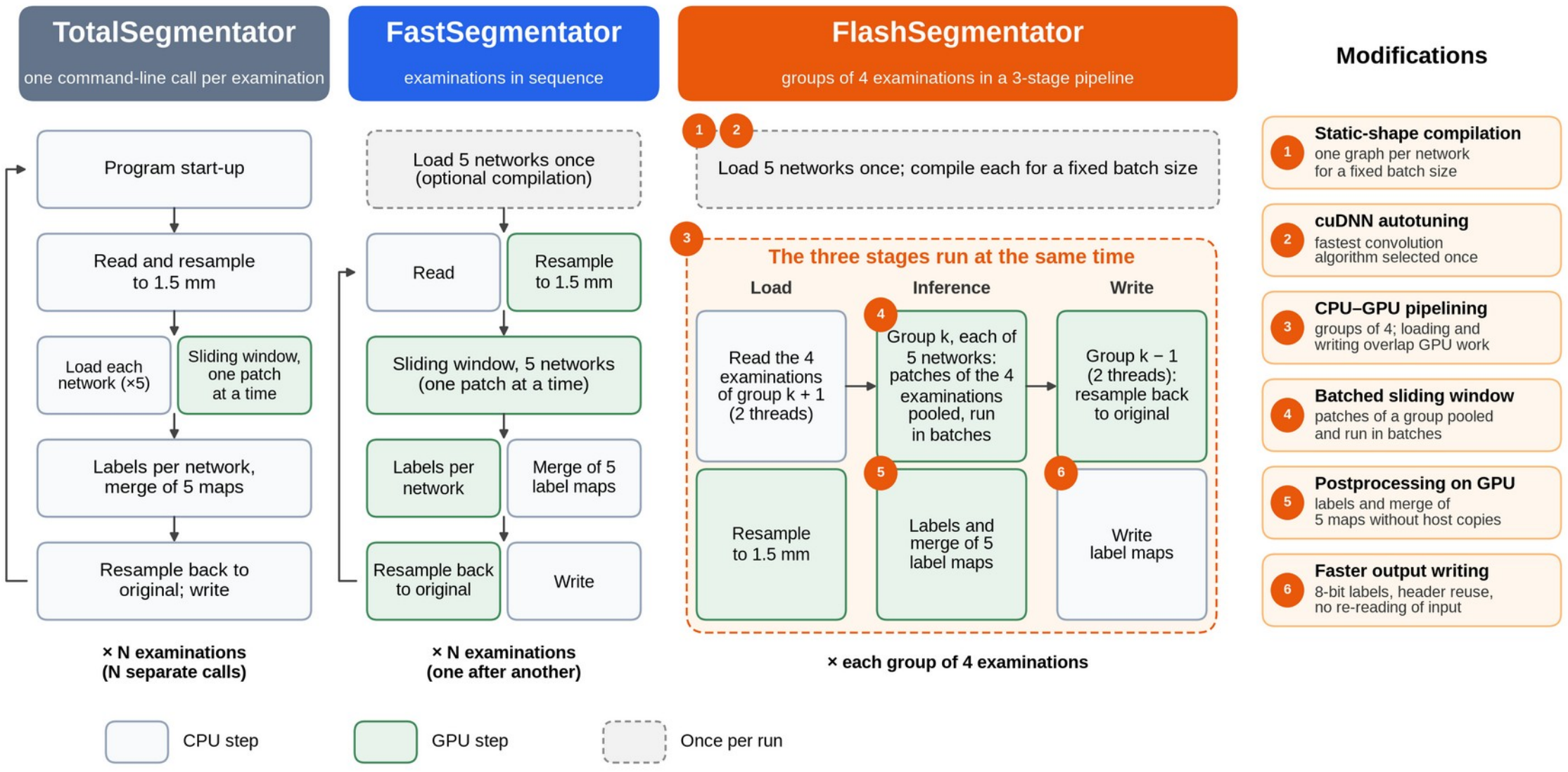

*Figure 1: Processing flow of the reference implementation (TotalSegmentator command-line interface), FastSegmentator, and FlashSegmentator for the CT total task. White boxes indicate CPU steps, green boxes GPU steps, and dashed gray boxes steps performed once per run. FlashSegmentator processes the examinations in groups of four in a three-stage pipeline, in which loading of group k + 1, inference on group k, and resampling back and writing of group k − 1 run at the same time. Circled numbers correspond to the modifications listed in Materials and Methods. MRI follows the same flow with two networks.*

## Hardware and software

Three GPU environments were used: an NVIDIA GH200 Grace Hopper Superchip in a node of the Miyabi-G supercomputer (Information Technology Center, The University of Tokyo) and NVIDIA A100 and T4 GPUs in Microsoft Azure virtual machines (Japan East region), all with PyTorch 2.11.0 and CUDA 12.8. One GPU was used for every run, and all implementations on a given machine shared the same weight files. GPU and CPU models, memory, software versions, and the conditions evaluated on each platform are listed in Table S1.

## Timing

Throughput was the number of examinations divided by the wall-clock time from the start of the first to the completion of the last examination, including input reading and output writing. One warm-up examination was processed before timing, and model loading and one-time compilation of FastSegmentator and FlashSegmentator were excluded; reference times included start-up and model loading of each call.

## Concurrent processes and cost

Because a single process may not fully use a GPU, each implementation was also run as concurrent processes on one GPU, using the 512 training-split examinations on the GH200 and A100 (the reference implementation on the A100 only, because concurrent calls exceeded the process limit of the GH200 node). For FastSegmentator and FlashSegmentator, the examinations were divided among K processes, and throughput was the total number of examinations divided by the time of the longest-running process. Processing cost per 10,000 examinations on Azure virtual machines was calculated from throughput and the pay-as-you-go list price.

## Application to MRI

On the A100, the reference implementation and FlashSegmentator (batch size 4, without body cropping) were compared on the 55 test MRI examinations.

## Statistical analysis

Although designed to be mathematically equivalent, outputs could differ slightly owing to floating-point arithmetic order and resampling; agreement was therefore tested. The Dice similarity coefficient was calculated for each structure present in the reference standard, and per-examination means were averaged. Agreement between implementations was assessed with paired differences of Dice at the examination level and the structure level (structures present in at least three examinations) using two one-sided tests (TOST) (18) with an equivalence margin of ±0.01 (90% CI within the margin). Body cropping was assessed for noninferiority with the same margin. Analyses used Python 3 with NumPy and SciPy.

# Results

## Dataset characteristics

The 16 excluded examinations (see Materials and Methods) had a shorter scan range (median, 288 vs 429 mm) and included three head CT examinations; FlashSegmentator, which reuses the input header, processed all 16 (mean Dice, 0.9452). All 30 BTCV examinations were processed by every implementation. The 55 MRI test examinations covered various body regions and sequences (spin-echo and gradient-echo; 1.5 and 3 T), with thicker sections and smaller images than the CT examinations (Table S2).

## Throughput

FlashSegmentator processed 3782, 1810, and 270 examinations per hour on the GH200, A100, and T4, respectively: 57.7, 33.9, and 6.4 times the reference implementation and 3.20, 2.50, and 1.22 times FastSegmentator with compilation (Table 1). With body cropping, throughput increased to 4695, 2374, and 362 examinations per hour.

**Table 1: Throughput on Three GPUs (TotalSegmentator Test Set, n = 73)**

| GPU | Implementation | Time per examination (s) | Examinations per hour | Relative throughput |
|---|---|---|---|---|
| GH200 | TotalSegmentator | 55.0 | 65.5 | 1.0 |
| | FastSegmentator | 3.26 | 1104 | 16.9 |
| | FastSegmentator, compiled | 3.05 | 1182 | 18.0 |
| | FlashSegmentator (B = 8) | 0.95 | 3782 | 57.7 |
| | FlashSegmentator (B = 8), body cropping† | 0.77 | 4695 | 71.7 |
| A100 | TotalSegmentator | 67.4 | 53.4 | 1.0 |
| | FastSegmentator | 5.22 | 689 | 12.9 |
| | FastSegmentator, compiled | 4.97 | 724 | 13.6 |
| | FlashSegmentator (B = 4) | 1.99 | 1810 | 33.9 |
| | FlashSegmentator (B = 4), body cropping† | 1.52 | 2374 | 44.5 |
| T4 | TotalSegmentator | 84.8 | 42.5 | 1.0 |
| | FastSegmentator | 18.0 | 200 | 4.7 |
| | FastSegmentator, compiled | 16.3 | 221 | 5.2 |
| | FlashSegmentator (B = 4) | 13.3 | 270 | 6.4 |
| | FlashSegmentator (B = 2), body cropping† | 9.95 | 362 | 8.5 |

Relative throughput is relative to the reference implementation on the same GPU. Reference times include start-up and model loading of each command-line call. † Optional body cropping (Table S5); on the GH200 measured in a separate run in which FlashSegmentator without cropping processed 3723 examinations per hour.

## Concurrent processes

On the 512 examinations, FlashSegmentator was fastest with a single process (GH200, 4707 examinations per hour; A100, 2263 per hour), and additional processes decreased its throughput (Figure 2, Table S3). Compiled FastSegmentator was fastest with four processes (GH200, 3079 per hour; A100, 1707 per hour), and the reference implementation reached 482 examinations per hour with 32 concurrent calls on the A100. Mean GPU utilization of the reference implementation, compiled FastSegmentator, and FlashSegmentator was 7%, 48%, and 93% with a single process and 45%, 97%, and 99% with multiple processes (Figure 3). With the best configuration of each implementation, FlashSegmentator was 1.53 (GH200) and 1.33 (A100) times faster than compiled FastSegmentator and 4.7 times faster than the reference implementation on the A100.

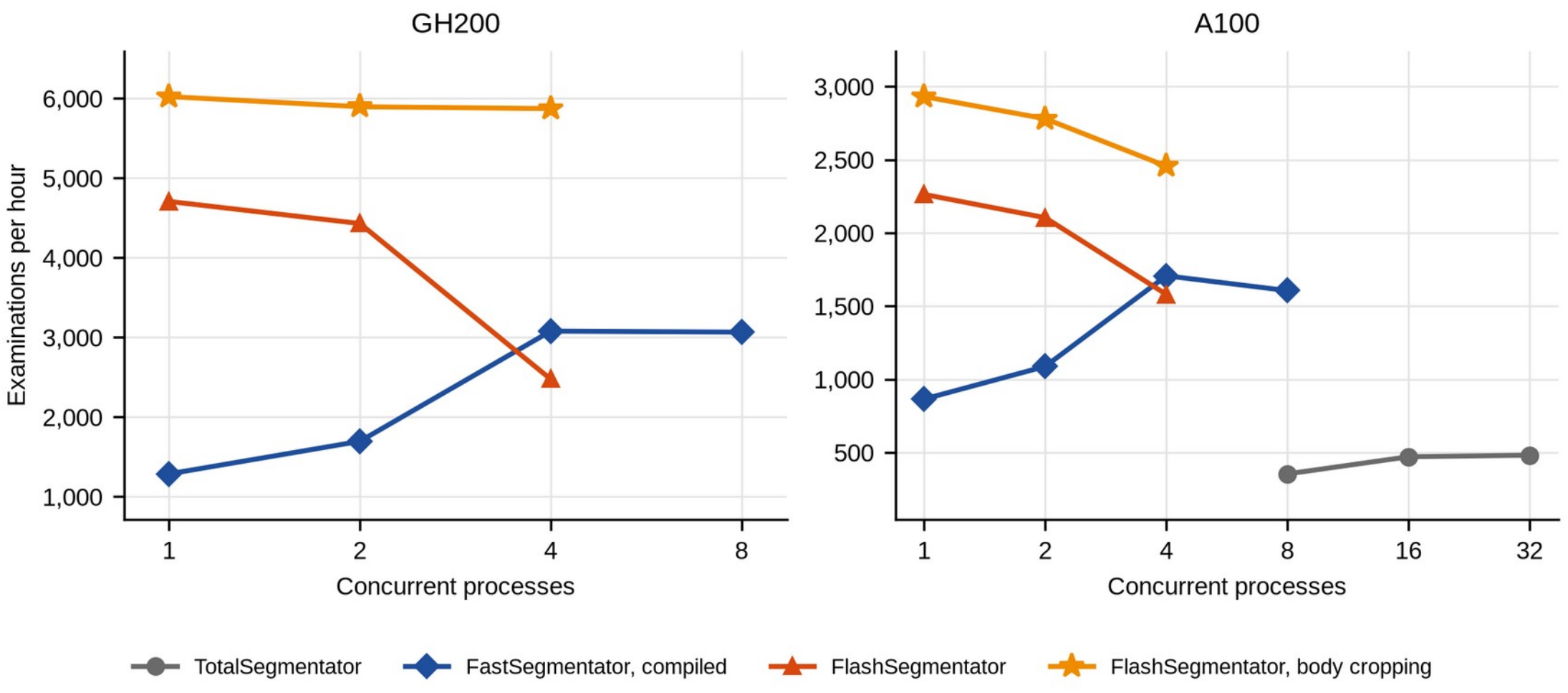

*Figure 2: Aggregate throughput with concurrent processes on one GPU (512 examinations from the training split of the TotalSegmentator dataset). The reference implementation was run as N concurrent command-line calls, and FastSegmentator (compiled) and FlashSegmentator as K concurrent processes; throughput is the number of examinations divided by the processing time of the longest-running process. Configurations that ran out of GPU memory are not shown. Numerical values, including out-of-memory configurations and smaller batch sizes for FlashSegmentator, are given in Table S3.*

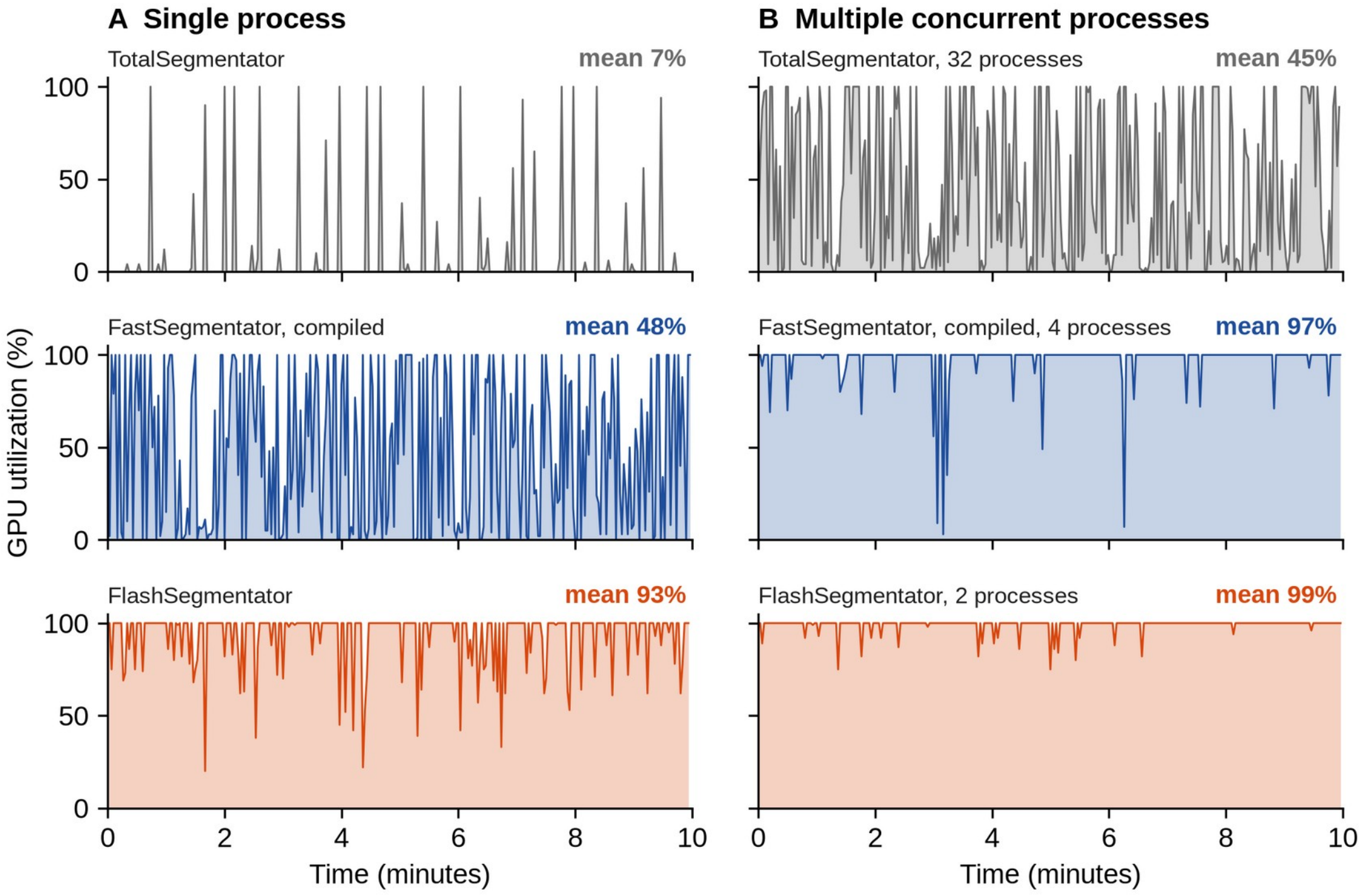

*Figure 3: GPU utilization during processing of the 512 training-split examinations on the A100 (10 minutes, starting 2 minutes after the first GPU activity), recorded with nvidia-smi at intervals of approximately 2 seconds. (A) One process; (B) multiple concurrent processes (reference implementation, 32 command-line calls; compiled FastSegmentator, four processes; FlashSegmentator, two processes). Rows show, from top, the reference implementation (command-line calls), compiled FastSegmentator,*

*and FlashSegmentator (batch size 4). The mean over the 10 minutes is shown at the upper right of each panel.*

## Segmentation accuracy

FlashSegmentator was equivalent to the reference implementation on the GH200, A100, and T4, at the examination level and for all 117 structures (Table 2); on the GH200, mean Dice was 0.93765 for the reference implementation, 0.93765 for FastSegmentator, and 0.93766 for FlashSegmentator. The widest structure-level 90% CI bound was 0.0021. Of the 5329 examination–structure pairs on the GH200, the largest absolute Dice difference was 0.026 (99th percentile, 0.0004). On BTCV, whose images are resampled to 1.5 mm before inference, FlashSegmentator was equivalent to the reference implementation at the examination level (mean difference, −0.0015; 90% CI: −0.0026, −0.0004); 12 of 13 structures were equivalent, and the portal and splenic veins narrowly missed the margin (lower 90% CI bound, −0.0105).

**Table 2: Equivalence of Dice Scores between Implementations (TOST, Margin ±0.01)**

| Dataset (GPU) | Comparison | n | Mean difference [90% CI] | Structures equivalent | Widest structure CI bound |
|---|---|---|---|---|---|
| TS test (GH200) | FastSegmentator vs reference | 73 | $1.3 \times 10^{-6}$ [$-1.7 \times 10^{-6}$, $4.4 \times 10^{-6}$] | 117/117 | 0.0006 |
| TS test (GH200) | FlashSegmentator vs reference | 73 | $9.7 \times 10^{-6}$ [$-2.8 \times 10^{-6}$, $2.2 \times 10^{-5}$] | 117/117 | 0.0021 |
| TS test (A100) | FlashSegmentator vs reference | 73 | $1.1 \times 10^{-5}$ [$8.3 \times 10^{-7}$, $2.1 \times 10^{-5}$] | 117/117 | 0.0006 |
| TS test (T4) | FlashSegmentator vs reference | 73 | $1.2 \times 10^{-5}$ [$1.8 \times 10^{-6}$, $2.3 \times 10^{-5}$] | 117/117 | 0.0013 |
| BTCV (GH200) | FlashSegmentator vs reference | 30 | $-1.5 \times 10^{-3}$ [$-2.6 \times 10^{-3}$, $-4.2 \times 10^{-4}$] | 12/13 | 0.0105 |

Body cropping increased throughput by a factor of 1.26–1.37 and was noninferior to the reference implementation at the examination level on each GPU (difference, +0.0011; 90% CI: 0.0002, 0.0021), with 106 of 117 structures noninferior (Table S5). On BTCV, it did not change Dice compared with FlashSegmentator without cropping (difference, +0.0001; 90% CI: −0.0009, 0.0011).

## Comparison with the lower-resolution modes

The --fast and --fastest modes of the reference implementation processed 181 and 213 examinations per hour on the A100, compared with 53.4 for the standard mode and 1810 for FlashSegmentator (Table 3). They reduced mean Dice by 0.116 and 0.241, most for thin or small structures (decrease of more than 0.2 in 12 and 74 structures; for example, the sixth cervical vertebra, 0.944 to 0.600 and 0.444), whereas the mean Dice of FlashSegmentator differed from the standard mode by 0.00001. In an example examination, the lower-resolution modes produced coarse, stair-stepped surfaces and fragmented ribs, whereas the output of FlashSegmentator preserved the same structures as that of the standard mode (Figure 4).

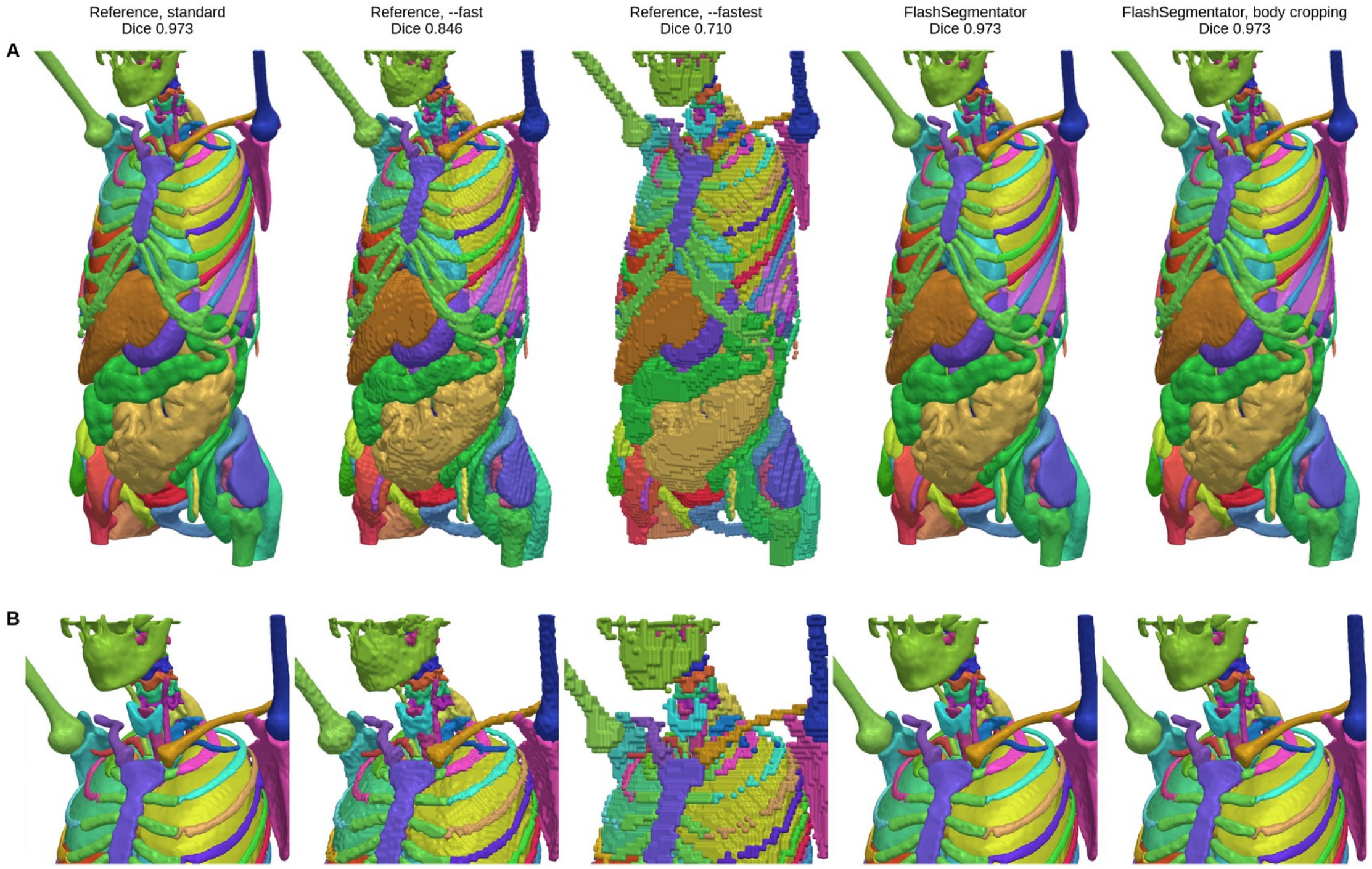

*Figure 4: Three-dimensional surface renderings of the segmentation of one TotalSegmentator test examination (114 structures in the reference standard) by the reference implementation in the standard, --fast, and --fastest modes and by FlashSegmentator without and with body cropping. (A) Anterior oblique view of the whole scan range. (B) Close-up of the neck and upper thorax. Each structure has the same color in all panels. Dice is the mean over the structures in the reference standard of this examination.*

**Table 3: Throughput and Mean Dice by Structure Group for the Lower-Resolution Modes (TotalSegmentator Test Set, n = 73)**

| Structure group | Structures | TotalSegmentator, standard | TotalSegmentator, --fast | TotalSegmentator, --fastest | FlashSegmentator (B = 4) |
|---|---|---|---|---|---|
| Examinations per hour | | 53 | 181 | 213 | 1810 |
| Mean Dice, all structures | 117 | 0.938 | 0.821 | 0.697 | 0.938 |
| Organs and others | 32 | 0.883 | 0.795 | 0.701 | 0.883 |
| Vertebrae | 25 | 0.952 | 0.806 | 0.678 | 0.952 |
| Ribs | 24 | 0.954 | 0.765 | 0.577 | 0.954 |
| Vessels | 13 | 0.917 | 0.786 | 0.614 | 0.917 |
| Muscles | 10 | 0.972 | 0.915 | 0.856 | 0.972 |

All values were obtained on the A100 (batch size 4 for FlashSegmentator); as in Table 1, each command-line call of the reference implementation included start-up and model loading. Structures were grouped by name; the remaining 13 structures (bones of the limbs and girdles, sacrum, sternum, and skull) were not grouped.

## Contribution of each modification

Static-shape compilation gave the largest gain (Table 4); it reduced the forward time per patch per network at batch size 8 from 8.10 to 5.55 ms.

**Table 4: Cumulative Effect of Each Modification on the GH200 (TotalSegmentator Test Set, n = 73)**

| Configuration | Batch size | Examinations per hour | Relative to FastSegmentator |
|---|---|---|---|
| FastSegmentator | 1 | 1104 | 1.00 |
| + Batched sliding window and pipelining | 4 | 1670 | 1.51 |
| + GPU postprocessing and faster output writing | 4 | 2283 | 2.07 |
| + Static-shape compilation | 2 | 3541 | 3.21 |
| + cuDNN autotuning, batch size 8 (final) | 8 | 3782 | 3.43 |

## Effect of batch size

Throughput varied by less than 7% across batch sizes on every GPU, whereas peak GPU memory increased up to twofold with batch size (Figure S1). The highest throughput was obtained at a batch size of 8 on the GH200 and 4 on the A100 and T4.

## Comparison of inference backends and network configurations

TensorRT, ONNX Runtime, fusion of the five networks into one grouped-convolution network, and a channels-last memory layout did not shorten the forward pass compared with the compiled PyTorch network (Table S6) and were not adopted.

## Processing cost on cloud virtual machines

Per 10,000 examinations, the processing cost on the A100 virtual machine was US $997 for the reference implementation run one call at a time (US $110 with 32 concurrent calls) and US $29 for FlashSegmentator (US $22 with body cropping); on the T4 virtual machine, it was US $167, US $26, and US $20 (Figure 5, Table S4).

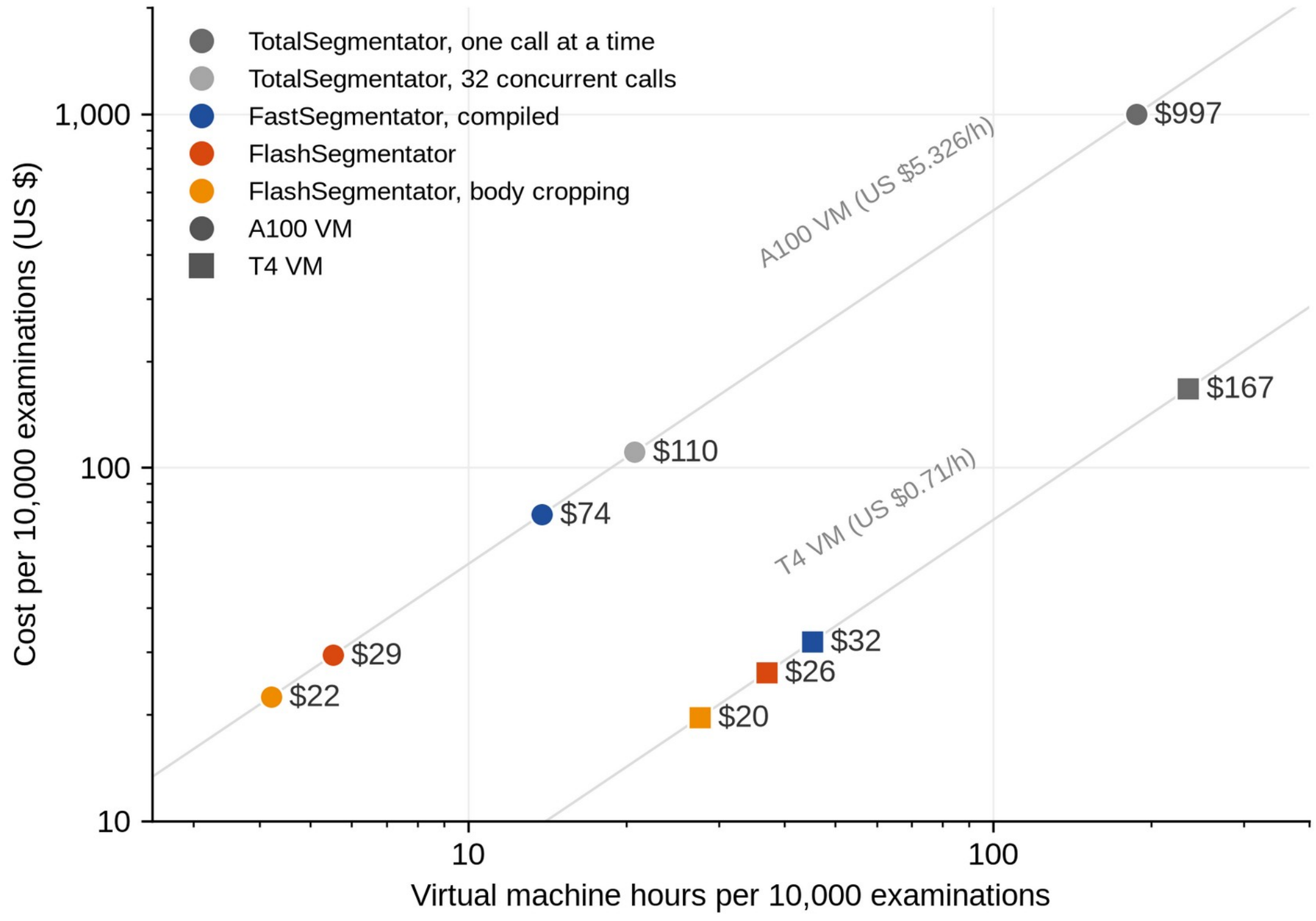

*Figure 5: Estimated processing time and cost per 10,000 examinations on Azure virtual machines (both axes logarithmic). Cost is the virtual machine time multiplied by the pay-as-you-go list price on November 18, 2026 (Linux, Japan East: NC24ads_A100_v4 for the A100, US $5.326 per hour; NC4as_T4_v3 for the T4, US $0.71 per hour), excluding storage and data transfer; the gray lines connect points of equal hourly price. Throughput was measured on the TotalSegmentator test set (Table 1), except for 32 concurrent calls of the reference implementation, which was measured on the 512 training-split examinations (Table S3). Numerical values are given in Table S4.*

## Throughput and segmentation agreement on MRI

On the A100, FlashSegmentator processed the 55 test MRI examinations in 0.27 seconds per examination, compared with 25.8 seconds for the reference implementation, a 94-fold increase in throughput (Table S7). Mean Dice was 0.78157 for the reference implementation and 0.78158 for FlashSegmentator; the per-examination difference was $1.4 \times 10^{-5}$ (90% CI: $-1.0 \times 10^{-5}$, $3.8 \times 10^{-5}$; TOST $P < .001$), and all 50 structures met the equivalence criterion (widest structure-level 90% CI bound, 0.0010).

# Discussion

In this study, optimization of the inference pipeline increased the batch-processing throughput of CT examinations while retaining the trained weights and standard resolution of TotalSegmentator. FlashSegmentator processed approximately 3800 examinations per hour on the GH200, and with its default settings its Dice scores were equivalent to those of the reference implementation. These results indicate that, even when an existing model continues to be used, revising its inference implementation can shorten the time needed to process large CT datasets.

A previous study reduced the computation of TotalSegmentator by Tucker decomposition of its convolution kernels and assessed the network forward time (19); in contrast, the present study kept the model unchanged and assessed throughput per examination, from reading the image to writing the segmentation.

Running several processes on one GPU is a common way to increase throughput, but it requires user-side preparation and did not close the gap. The reference implementation on the A100 reached only 482 examinations per hour with 32 concurrent calls, compared with 471 with 16, indicating that parallelization alone has a ceiling and that the inference itself needs to be made more efficient. FlashSegmentator overlaps CPU and GPU work within one process and already kept the GPU busy with a single process, so that additional processes competed for GPU memory. For larger workloads, assigning one FlashSegmentator process to each GPU is therefore likely to scale better than running several processes on one GPU.

The lower-resolution modes of the reference implementation increase throughput, but they substantially reduced Dice scores for structures such as ribs and some vertebrae. FlashSegmentator retained the accuracy of the standard mode and was still approximately 8 to 10 times faster than the lower-resolution modes on the A100. Dice scores were equivalent on the TotalSegmentator test set and, at the examination level, on BTCV, an external benchmark with manual reference segmentations, where 12 of 13 structures met the equivalence criterion. Because the optional body cropping was not noninferior for all structures, the default settings without cropping appear appropriate for studies that quantify the whole body at standard resolution, including thin and small structures.

In large-scale CT studies, automated segmentation has been applied to datasets of tens of thousands to more than 350,000 examinations. Reference curves for whole-body anatomic volumes were derived from approximately 190,000 scans in an institutional PACS and two external datasets (7), and whole-body attenuation and volume charts were constructed from more than 350,000 CT examinations after filtering pathologic findings from the reports with large language models (6). In a study of abdominal organ volumes using the Japan Medical Image Database (J-MID), a nationwide Japanese repository of medical images, the lower-resolution mode of TotalSegmentator was used to process approximately 50,000 abdominopelvic CT examinations on a single GPU (8). A cardiac quantification study in the NLST included approximately 24,000 participants (20), and approximately 130,000 CT series were processed in a distributed manner in the cloud in the NCI Imaging Data Commons (21). On the basis of the measured throughput of FlashSegmentator, segmentation of 100,000 CT examinations at standard resolution would take approximately 26 hours on a single node with one GH200. If the examinations were divided equally among eight such nodes and each node maintained the same throughput, the processing time would be shortened to approximately 3.3 hours. These estimates provide a guide to the segmentation time required for the initial analysis of a large cohort or for reanalysis after a model update.

## Limitations

In this study, the implementations were compared on several GPU environments. Each condition was timed once in principle, and run-to-run variation under identical conditions was not assessed. Concurrent processes were evaluated on 512 training-split examinations on the GH200 and A100 only; on GPUs with less memory, such as the T4, the configurations that can be run concurrently are limited. For images not already at 1.5-mm resolution, small differences from the reference implementation can arise from the implementation of resampling, as observed for one structure on BTCV. The processing time for 100,000 examinations is an estimate that assumes the measured throughput is sustained; the actual time will depend on the computing environment, storage performance, and the time needed for image retrieval from the PACS and DICOM conversion. In addition, the evaluation was limited to the CT total and MRI total_mr tasks of TotalSegmentator, and the extent to which the same optimizations are effective for organ-specific tasks or other segmentation models remains to be investigated.

## Conclusion

FlashSegmentator increased batch-processing throughput while retaining the trained weights and standard resolution of TotalSegmentator, with Dice scores equivalent to those of the reference implementation, and can shorten the computation time for the analysis of large CT cohorts.

# Acknowledgments

Part of the computations were performed on the Miyabi-G supercomputer at the Joint Center for Advanced High Performance Computing (Information Technology Center, The University of Tokyo, and Center for Computational Sciences, University of Tsukuba). In this research work, we used the UTokyo Azure (https://utelecon.adm.u-tokyo.ac.jp/en/research_computing/utokyo_azure/). UTokyo Azure was used only for the measurements on the NVIDIA A100 and NVIDIA T4 GPUs; all other experiments were performed on other computing resources.

# Funding

This work was supported by JST, CREST Grant Number JPMJCR21M2, Japan.

# Code and Data Availability

The source code of FlashSegmentator, together with the scripts used for the throughput and agreement analyses, is available at https://github.com/yamagishi0824/FlashSegmentator under the Apache License 2.0. All datasets used in this study are publicly available: the TotalSegmentator CT and MRI datasets on Zenodo (13,16) and the BTCV dataset (14).

# Supplemental Material

**Body cropping.** Before sliding-window inference, the resampled 1.5-mm image was cropped to the bounding box of the body (voxels above −500 HU after morphologic opening, components of at least 1% of the largest component, 10-voxel margin), and the segmentation was pasted back into a background volume of the original size. Because voxels outside the box are always labeled background and the patch grid changes, the output differs slightly from that of the reference implementation. Body cropping was not used for MRI.

### Table S1: Hardware, Software, and Inference Settings of Each Platform

| Item | GH200 | A100 | T4 |
|---|---|---|---|
| Platform | Miyabi-G supercomputer node (Information Technology Center, The University of Tokyo) | Microsoft Azure virtual machine, NC24ads_A100_v4 (Japan East) | Microsoft Azure virtual machine, NC4as_T4_v3 (Japan East) |
| GPU | NVIDIA GH200 Grace Hopper Superchip (H100-class GPU) | NVIDIA A100 PCIe | NVIDIA T4 |
| GPU memory | 96 GB HBM3 | 80 GB HBM2e | 16 GB GDDR6 |
| CPU | NVIDIA Grace, 72 Arm Neoverse V2 cores | AMD EPYC 7V13, 24 vCPUs | AMD EPYC 7V12, 4 vCPUs |
| System memory | 120 GB LPDDR5X | 220 GiB | 28 GiB |
| NVIDIA driver | 615.71.09 | 580.178.04 | 580.178.04 |
| Python | 3.11.15 | 3.11.18 | 3.11.18 |
| PyTorch (CUDA) | 2.11.0 (12.8) | 2.11.0 (12.8) | 2.11.0 (12.8) |
| Data sets | TS test (73), BTCV (30), TS train (512) | TS test (73), TS MRI test (55), TS train (512) | TS test (73) |
| Implementations | Reference, FastSegmentator (± compilation), FlashSegmentator | Reference (standard, --fast, --fastest), FastSegmentator (± compilation), FlashSegmentator | Reference, FastSegmentator (± compilation), FlashSegmentator |
| FlashSegmentator batch size (main analysis) | 8 | 4 | 4 (2 with body cropping) |
| Batch sizes evaluated | 2, 4, 8, 16 | 2, 4, 8 | 1, 2, 4 |
| Concurrent processes evaluated | Others 1–16 (reference not evaluated) | Reference 8, 16, 32 calls; others 1, 4, 8 | Not evaluated |
| Reference timing | Per command-line call, including start-up and model loading | Same as GH200 | Same as GH200 |

Reference implementation: TotalSegmentator 2.18.0 (commit 8806051) with nnunetv2 2.8.1. FastSegmentator: commit c8040ff. FastSegmentator and FlashSegmentator: the FastSegmentator fork of nnunetv2 (2.5.2). GPU memory is the vendor specification (the GH200 GPU reported 97,871 MiB). CPU models of the Azure virtual machines are those listed by Microsoft for the virtual machine series. TS = TotalSegmentator.

## Table S2: Characteristics of the Datasets

| Characteristic | TotalSegmentator test (n = 73) | TotalSegmentator train, concurrent-process analysis (n = 512) | BTCV (n = 30) | TotalSegmentator MRI test (n = 55) |
|---|---|---|---|---|
| Age (y)* | 65 (52–74) | 65 (56–76) | NA | 57 (45–67) |
| Sex, male/female/unknown | 37/35/1 | 297/214/1 | NA | 20/19/16 |
| Manufacturer, Siemens/GE/Philips/unknown | 45/9/6/13 | 451/25/36/0 | NA | 34/14/5/2 |
| Magnetic field strength, 1.5 T/3 T/unknown | NA | NA | NA | 28/25/2 |
| Voxel spacing (mm) | 1.5 isotropic | 1.5 isotropic | 0.6–1.0 in-plane, 2.5–5.0 section | 0.8 (0.5–1.2) in-plane, 3.3 (3.0–4.8) section* |
| Number of sections* | 286 (219–370) | 240 (156–393) | 128 (97–148) | 36 (26–75) |
| Scan range (mm)* | 429 (329–555) | 361 (234–590) | 445 (421–490) | 216 (160–278) |
| Image size (million voxels)* | 17.7 (11.7–28.1) | 13.5 (6.4–24.2) | 33.4 (25.3–38.7) | 6.7 (3.3–9.3) |
| Reference standard structures | 117 | Not used | 13 | 50 |

* Median (interquartile range); age missing for one TotalSegmentator test CT examination and 16 MRI examinations. The training-split examinations were used only for the analysis of concurrent processes. For MRI, section thickness is the largest voxel dimension and scan range is the craniocaudal extent. TotalSegmentator images are distributed resampled to 1.5 mm. NA = not available.

**Table S3: Aggregate Throughput (Examinations per Hour) with Concurrent Processes on One GPU (512 Training-Split Examinations)**

| GPU | Implementation | 1 | 2 | 4 | 8 | 16 | 32 |
|---|---|---|---|---|---|---|---|
| GH200 | TotalSegmentator | NE | NE | NE | NE | NE | NE |
| | FastSegmentator, compiled | 1287 | 1696 | 3079 | 3067 | OOM | — |
| | FlashSegmentator (B = 8) | 4707 | 4430 | 2476 | OOM | OOM | — |
| | FlashSegmentator (B = 4) | — | — | 4444 | — | — | — |
| | FlashSegmentator (B = 2) | — | — | 4260 | OOM | — | — |
| | FlashSegmentator (B = 1) | — | — | — | OOM | — | — |
| | FlashSegmentator, body cropping (B = 8) | 6023 | 5896 | 5872 | OOM | — | — |
| A100 | TotalSegmentator | — | — | — | 354 | 471 | 482 |
| | FastSegmentator, compiled | 865 | 1088 | 1707 | 1606 | — | — |
| | FlashSegmentator (B = 4) | 2263 | 2104 | 1580 | — | — | — |
| | FlashSegmentator (B = 2) | — | — | 2096 | — | — | — |
| | FlashSegmentator, body cropping (B = 4) | 2931 | 2780 | 2456 | — | — | — |

Columns give the number of concurrent processes (reference implementation: concurrent command-line calls). Throughput is 512 examinations divided by the processing time of the longest-running process (reference implementation: total wall-clock time). OOM = at least one process ran out of GPU memory. NE = not evaluated (on the GH200 node, 16 or more concurrent calls of the reference implementation failed because of multiprocessing errors). — = not measured. B = batch size.

### Table S4: Estimated Processing Time and Cost per 10,000 Examinations on Azure Virtual Machines

| Virtual machine (list price) | Configuration | Examinations per hour | Virtual machine hours | Cost (US $) |
|---|---|---|---|---|
| A100 (US $5.326/h) | TotalSegmentator, one call at a time | 53.4 | 187.3 | 997 |
| | TotalSegmentator, 32 concurrent calls‡ | 482 | 20.7 | 110 |
| | FastSegmentator | 689 | 14.5 | 77 |
| | FastSegmentator, compiled | 724 | 13.8 | 74 |
| | FastSegmentator, compiled, 4 processes‡ | 1707 | 5.9 | 31 |
| | FlashSegmentator | 1810 | 5.5 | 29 |
| | FlashSegmentator, body cropping | 2374 | 4.2 | 22 |
| T4 (US $0.71/h) | TotalSegmentator, one call at a time | 42.5 | 235.3 | 167 |
| | FastSegmentator | 200 | 50.0 | 36 |
| | FastSegmentator, compiled | 221 | 45.2 | 32 |
| | FlashSegmentator | 270 | 37.0 | 26 |
| | FlashSegmentator, body cropping | 362 | 27.6 | 20 |

Calculated from the measured throughput (Table 1 and Table S3) and the pay-as-you-go list price on November 18, 2026 (Linux, Japan East: NC24ads_A100_v4 for the A100, NC4as_T4_v3 for the T4); storage and data transfer excluded. ‡ Measured on the 512 training-split examinations (Table S3); other rows, TotalSegmentator test set (Table 1).

### Table S5: Throughput and Noninferiority of FlashSegmentator with Body Cropping (TotalSegmentator Test Set, n = 73)

| GPU | Without cropping (per hour) | With cropping (per hour) | Ratio | Dice with cropping | Difference vs reference [90% CI] | Structures noninferior |
|---|---|---|---|---|---|---|
| GH200 (B = 8) | 3723 | 4695 | 1.26 | 0.93880 | +0.0012 [0.0002, 0.0021] | 106/117 |
| A100 (B = 4) | 1810 | 2374 | 1.31 | 0.93878 | +0.0011 [0.0002, 0.0021] | 106/117 |
| T4 (B = 2) | 265 | 362 | 1.37 | 0.93878 | +0.0011 [0.0002, 0.0021] | 106/117 |

Noninferiority margin −0.01 Dice. Reference implementation mean Dice, 0.93765.

### Table S6: Network Forward Time by Inference Backend and Network Configuration (GH200)

| Configuration | Batch size 1 | Batch size 8 |
|---|---|---|
| PyTorch, compiled (adopted) | 6.81 | 5.55 |
| PyTorch, not compiled | 13.40 | 8.10 |
| TensorRT, FP16 | 10.7 | 9.2 |
| TensorRT, FP32 | 14.4 | 12.5 |
| ONNX Runtime (CUDA) | 16.4 | 15.2 |
| Five networks fused (grouped convolution), compiled | 8.24 | NA |
| Channels-last memory layout, compiled | NA | 5.62 |

Values are milliseconds per patch per network. For the fused network, the time for all five networks (41.2 ms, vs 33.7 ms for the five separate compiled networks) was divided by five. NA = not measured.

**Table S7: Throughput and Segmentation Agreement on MRI (A100, TotalSegmentator MRI Test Set)**

| Implementation | Time per examination (s) | Examinations per hour | Relative throughput | Mean Dice | Structures equivalent |
|---|---|---|---|---|---|
| TotalSegmentator | 25.8 | 140 | 1.0 | 0.78157 | — |
| FlashSegmentator (B = 4) | 0.27 | 13,139 | 94 | 0.78158 | 50/50 |

Throughput was measured on all 55 examinations; Dice was calculated on the 49 examinations with nonempty reference standard segmentations.

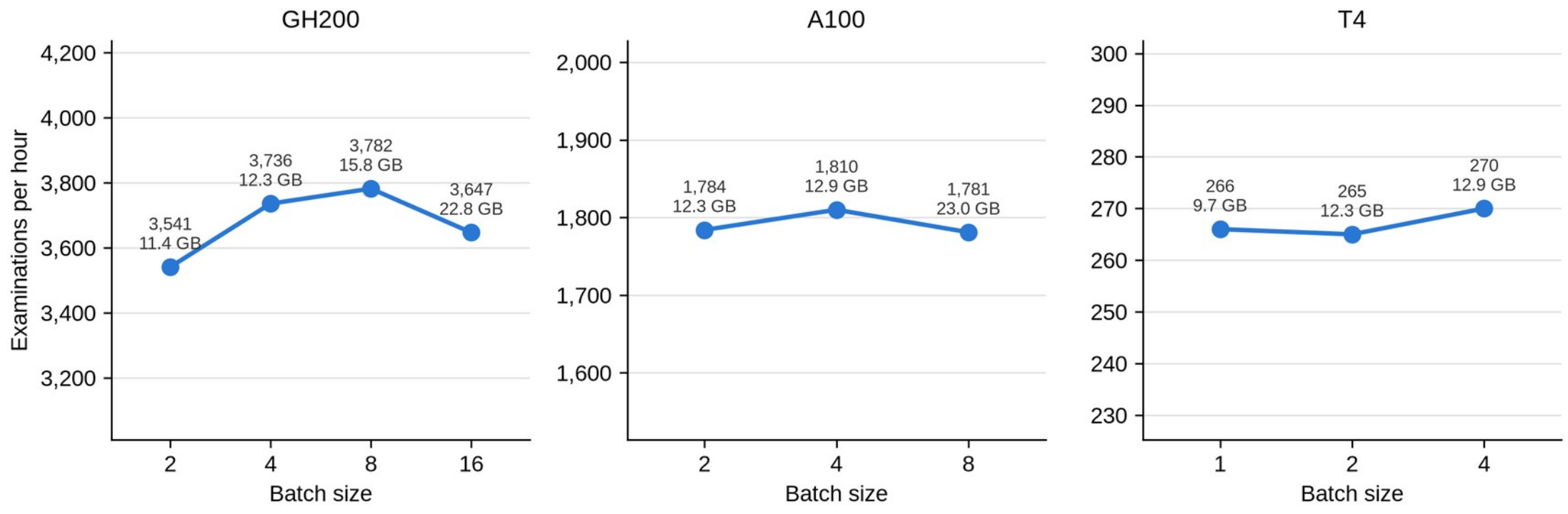

*Figure S1: Throughput of FlashSegmentator by batch size on each GPU (TotalSegmentator test set, n = 73). Labels show examinations per hour and peak GPU memory.*